\documentclass[preprint,12pt]{elsarticle}

\usepackage{amssymb}
\usepackage{multirow}
\usepackage{afterpage}
\usepackage{rotating}
\usepackage{amsmath}
\usepackage{tablefootnote}
\usepackage{graphicx} % Import the graphicx package
\usepackage{caption} % Import the caption package for captions
\usepackage{xurl}
\usepackage{soul}
\usepackage{xcolor} % For custom colors

\usepackage{ulem}
\usepackage{gensymb}
\usepackage{orcidlink}

\usepackage{setspace}
\usepackage[margin=1in]{geometry} % Set 1 inch margins by default

\usepackage{natbib} 

\journal{Annals of Biomedical Engineering}
       
\begin{document}

\begin{frontmatter}

% Define custom highlight color
\sethlcolor{yellow}

\title{Time-driven Survival Analysis from FDG-PET/CT in Non-Small Cell Lung Cancer}

\author[inst1]{Sambit Tarai\fnmark[*]\orcidlink{0000-0002-5550-3575}}
\author[inst1]{Ashish Chauhan\fnmark[*]\orcidlink{0009-0009-3050-1264}}
\author[inst1]{Elin Lundström\orcidlink{0000-0003-2955-4958}}
\author[inst1]{Johan Öfverstedt}
\author[inst3]{Therese Sjöholm\orcidlink{0000-0003-3072-749X}}
\author[inst3]{Veronica Sanchez Rodriguez}
\author[inst1,inst2]{Håkan Ahlström\orcidlink{0000-0002-8701-969X}}
\author[inst1,inst2]{Joel Kullberg\orcidlink{0000-0001-8205-7569}}

% Affiliations (same as before)
\affiliation[inst1]{organization={Radiology, Department of Surgical Sciences},
            addressline={Uppsala University},
            city={Uppsala},
            postcode={SE-75185},
            country={Sweden}}

\affiliation[inst2]{organization={Antaros Medical},
            city={Mölndal},
            postcode={SE-43153},
            country={Sweden}}

\affiliation[inst3]{organization={Molecular Imaging and Medical Physics, Department of Surgical Sciences},
            addressline={Uppsala University},
            city={Uppsala},
            postcode={SE-75185},
            country={Sweden}}

% Corresponding author + equal contribution note
%\cortext[cor]{Corresponding author}
%\footnotetext[1]{First authors.}
\fntext[*]{First authors.}

%\linenumbers
\begin{abstract}

\textbf{Purpose}: Automated medical image-based prediction of clinical outcomes, such as overall survival (OS), has great potential in improving patient prognostics and personalized treatment planning. We developed a deep regression framework using tissue-wise FDG-PET/CT projections as input, along with a temporal input representing a scalar time horizon (in days) to predict OS in patients with Non-Small Cell Lung Cancer (NSCLC). 

\textbf{Methods}: The proposed framework employed a ResNet-50 backbone to process input images and generate corresponding image embeddings. The embeddings were then combined with temporal data to produce OS probabilities as a function of time, effectively parameterizing the predictions based on time. The overall framework was developed using the U-CAN cohort (n = 556) and evaluated by comparing with a baseline method on the test set (n = 292). The baseline utilized the ResNet-50 architecture, processing only the images as input and providing OS predictions at pre-specified intervals, such as 2- or 5-year. 

\textbf{Results}: The incorporation of temporal data with image embeddings demonstrated an advantage in predicting OS, outperforming the baseline method with an improvement in ${AUC}$ of 4.3\%.
The proposed model using clinical + IDP features achieved strong performance, and an ensemble of imaging and clinical + IDP models achieved the best overall performance (0.788), highlighting the complementary value of multimodal inputs. The proposed method also enabled risk stratification of patients into distinct categories (high vs low risk). Heat maps from the saliency analysis highlighted tumor regions as key structures for the prediction. 

\textbf{Conclusion}: Our method provided an automated framework for predicting OS as a function of time, and demonstrates the potential of combining imaging and tabular data for improved survival prediction.

\end{abstract}


\end{frontmatter}

%\linenumbers
\section{Introduction}
\label{sec:sample1}

Non-Small Cell Lung Cancer (NSCLC) is one of the leading causes of cancer-related deaths globally \cite{me2024global} and despite innovations in treatment approaches, the 2- and 5-year overall survival (OS) remain low \cite{kolb2024molecular}. Accurate prediction of patient outcomes is highly valuable for informed decision-making, patient prognostics, and optimization of treatment plans. Current clinical practice in assessing patient outcome in NSCLC often involves the integration of clinical, pathological, and imaging data, with tumor-node-metastasis (TNM) staging \cite{lababede2018eighth} as a key determinant of OS. However, conventional prognostic models and scoring systems have limitations and may not fully leverage the rich information present in medical images \cite{alexander2017lung}. One such method is the Cox proportional hazards (CPH), a statistical method for survival analysis \cite{yang2019identifying}. 

\begin{figure}[!htb]
  \centering
  \includegraphics[width=1.00\textwidth]{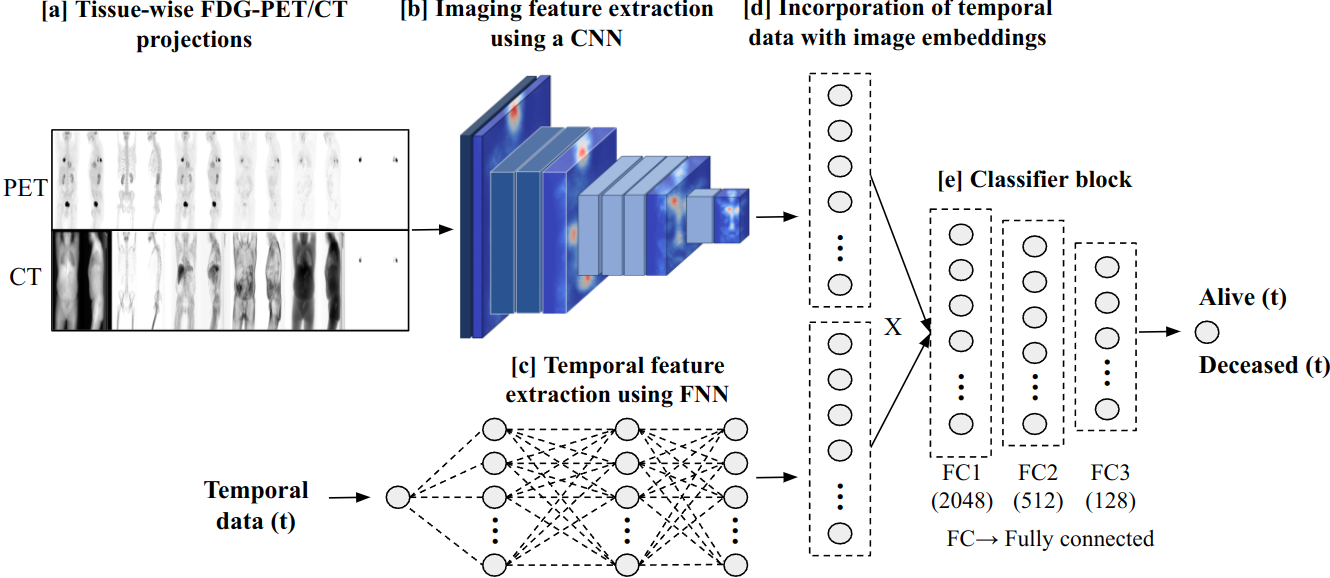}
  \caption{\small Overview of the proposed framework for automated OS prediction: [a] generation of tissue-wise PET/CT projections, [b] imaging feature extraction using a CNN, [c] temporal feature extraction using a FNN, [d] incorporation of temporal data with image embeddings, [e] classification of OS status for a given time.}
\end{figure}

Whole-body 18F-fluorodeoxyglucose-positron emission tomography/computed tomography (FDG-PET/CT) is extensively used in oncology for cancer \cite{almuhaideb201118f}. These images contain a wealth of information that may not be apparent through visual assessment. Convolutional neural networks (CNNs) have shown great potential to extract complex image features indicative of tumor progression \cite{oh2023deep} \cite{pedrosa2021lndb}. Previous studies have attempted the prediction of total metabolic tumor volume (TMTV) and Dmax which are recognized as significant predictors of OS \cite{girum202218f} \cite{tarai2024prediction}. Most of these studies have adopted a deep regression-based approach utilizing 2D projections rather than 3D volumes, to reduce computational costs and overfitting \cite{haggstrom2024deep}. Employing such methods has the potential to discover previously unknown associations between medical images and patient outcome and could hold great significance in clinical decision making.

CNN-based survival prediction has recently shown superior performance compared to traditional CPH-based approaches \cite{wiegrebe2024deep}. However, most existing methods frame OS prediction as a binary classification task at fixed time intervals (e.g., 2- or 5-year survival), which has several limitations. Patients lost to follow-up or not followed for the entire period, as well as those who die from unrelated causes, are often excluded, reducing the sample size and limiting the model's generalizability. Additionally, in traditional survival analysis, patients who die before a fixed time interval are treated equivalently, regardless of whether death occurs shortly after diagnosis or much later, overlooking important temporal dynamics. Such binary classification fails to capture the gradual changes in survival probability over time.

To address these limitations, we propose a time-driven deep learning (DL) framework that directly models OS probability as a continuous function of time, integrating tissue-wise FDG-PET/CT projections with temporal data defined as a scalar time horizon (in days) at which the patient’s OS probability is modelled. Unlike fixed-time classifiers, our approach dynamically adapts to evolving risk trajectories, offering a more nuanced and clinically relevant prediction of survival outcomes in NSCLC.

\section{Related works}

DL models have recently been used in survival analysis to enhance the predictive capabilities of traditional models. The CPH model, a widely used semi-parametric regression model for time-to-event predictions, has been extended through DL techniques. For instance, \cite{katzman2018deepsurv} introduced DeepSurv, which enhances the CPH model by replacing the linear predictor with a deep neural network to model non-linear interactions between covariates and survival outcomes. It demonstrated superior performance in terms of C-index but remained constrained by the proportionality assumption of the CPH model. To address this, \cite{kvamme2019time} proposed a model that incorporated time within the neural network, modelling time-dependent effects where the impact of risk factors changes over time. DeepHit \cite{lee2018deephit} is another DL-based survival analysis method that directly models the probability distribution of survival times without relying on the proportional hazards assumption. It effectively handles competing risks by jointly learning the distributions of different event types and has demonstrated strong performance in survival prediction tasks.
In addition, hybrid architectures combining convolutional and recurrent networks have been explored to capture temporal dependencies in survival prediction \cite{lu2023hybrid}. 
Recently, CNNs have gained popularity for survival analysis from medical images and have shown promising results in cancer survival prediction using both single modality and multimodal data \cite{wiegrebe2024deep}. A common drawback of these methods is that they often require pre-specified fixed-time survival analysis, e.g. prediction of 2- or 5-year survival rates. Our work distinguishes itself by combining the strengths of both approaches, utilizing feature embeddings from multimodal imaging data and temporal data to model survival probabilities as a continuous function of time. 
Unlike regression-based survival models that estimate relative risk (e.g., Cox-based approaches) or survival probabilities at predefined intervals (e.g., discrete-time models), the proposed framework directly conditions survival prediction on a continuous time horizon, enabling flexible estimation of time-dependent survival probabilities without requiring proportional hazards assumptions or fixed time discretization.

\section{Methods}

We propose a deep regression framework, incorporating the FDG-PET/CT projections and temporal data for the prediction of OS probabilities as a function of time. The overall end-to-end framework is illustrated in Figure 1.

\subsection{\textbf{Dataset (U-CAN)}}

This study utilized the NSCLC subset of the U-CAN cohort \cite{glimelius2018u}, comprising whole-body FDG-PET/CT images from 848 patients. Demographic and clinical characteristics of the cohort are detailed in Table A.3 in \ref{appendix:data_desc}. The voxel size in each of the FDG-PET/CT images was resampled to a uniform spacing of (2.04 x 2.04 x 3.00) mm³. Clinical variables such as age, sex, TNM staging, treatment type, OS time were collected. For our analysis, we included all patients with at least 90 days of follow-up and applied a maximum follow-up cut-off of 5 years. Automated tumor segmentation proposals were initially generated using a UNet-based method previously trained on the autoPET dataset \cite{tarai2024improved}. These segmentation proposals (n=556) were then reviewed and refined by a specialist in nuclear medicine and radiology (\textgreater 10 years of experience) to establish the ground truth (GT) tumor masks. The UNet-based method was subsequently fine-tuned using these GT masks to improve the quality of tumor segmentation proposals for the remaining cases, i.e., the test set (n=292). Ethical approval was obtained from the Swedish Ethical Review Authority (Dnr 2023-02312-02).

\subsection{\textbf{Proposed framework}}

\textbf{[a] Tissue-wise FDG-PET/CT projections:} The PET and CT volumes were categorized into specific tissue types: bone (if $i \geq 200$), lean soft tissue (if $-29 \leq i \leq 150$), adipose tissue (if $-190 \leq i \leq -30$), and air (if $i < -190$), where $i$ represents CT Hounsfield unit. 
This tissue-wise decomposition was introduced to capture physiologically distinct regions, enabling the model to learn heterogeneous metabolic and structural patterns (e.g., tumor burden in bone vs. soft tissue), which may have differential relevance for survival prediction.
Maximum and average intensity projections (MIPs and AIPs, respectively) were computed for all the PET and CT channels (denoted as $PET_{orig}^{MIP}$, $CT_{orig}^{AIP}$, $PET_{bone}^{MIP}$, $CT_{bone}^{AIP}$, etc) along the coronal and sagittal directions to obtain efficient 2D representations of the 3D volumes. Additionally, tumor segmentation masks were used to generate corresponding projections of tumor intensities only, denoted as $SEG^{MIP}$ and $SEG^{AIP}$. 

The resulting coronal and sagittal projections were resized using cropping or padding to a fixed size of $(400 \times 256)$, and then combined into a single image collage of size $(400 \times 512)$, resulting in 12-channel tissue-wise FDG-PET/CT projections (see Figure 1[a]). 
Detailed information can be found in \cite{tarai2025whole}.

\textbf{[b] Imaging feature extraction using a CNN backbone:} A ResNet-50 \cite{he2016deep} based CNN backbone was used as a feature extractor to process the 12-channel input images (see Figure 1[b]). The first convolutional layer was adapted to accommodate the increased number of input channels. The network generated robust image embeddings that capture the spatial, anatomical, and metabolic characteristics of the input projections. ResNet-50 was chosen as the feature extraction backbone for its computational efficiency, robust feature embeddings, and proven performance in medical image analysis.

\textbf{[c] Temporal feature extraction using FNN:} The temporal data were first sampled (in days) and then normalized to ensure they were on a consistent scale before being fed into a separate feed-forward neural network (FNN). This network projected the temporal data into the vector space of the image embeddings (see Figure 1[c]).

\textbf{[d] Incorporation of temporal data with image embeddings:} The image embeddings were then combined with the temporal data using element-wise multiplication (see Figure 1[d]) to predict time-specific OS probabilities. The temporal data served as a reference at which the patient's OS status was evaluated as either "deceased" or "alive".

\textbf{[e] Classifier block:} The classifier block processed the combined embeddings, which included features from image regions and temporal data, through a series of fully connected layers (see Figure 1[e]). It predicted the probability of a patient being "alive" or "deceased" at a given time-point. These time-specific probabilities were aggregated to construct a survival curve, which represented the OS probability as a function of time. The survival curve provided a continuous estimate of the likelihood of a patient surviving beyond a given time, enabling the prediction of OS.

\subsection{\textbf{Training strategy for the proposed framework}}

The proposed framework was trained to predict OS probabilities at any arbitrarily chosen time-points by sampling various time-points for each patient within the observation window. The observation window was the period from the scan date to the last follow-up date (LFD). 
The LFD corresponds to the date of the latest clinical data extract and represents the most recent time point at which survival information was available for the cohort; it defines the end of the observation period for patients who remained alive.
The dataset (n=848) was divided into a cross-validation set (n=556) with manual tumor segmentations and a test set (n=292) with automated tumor segmentations. In the cross-validation set (n=556), 328 patients survived (remained alive until the LFD), and 228 patients were deceased (died during the observation window). 

\begin{figure}[!htbp] % You can replace "htbp" with other options to control the image placement.
  \centering % Centers the image horizontally
  \includegraphics[width=0.95\textwidth]{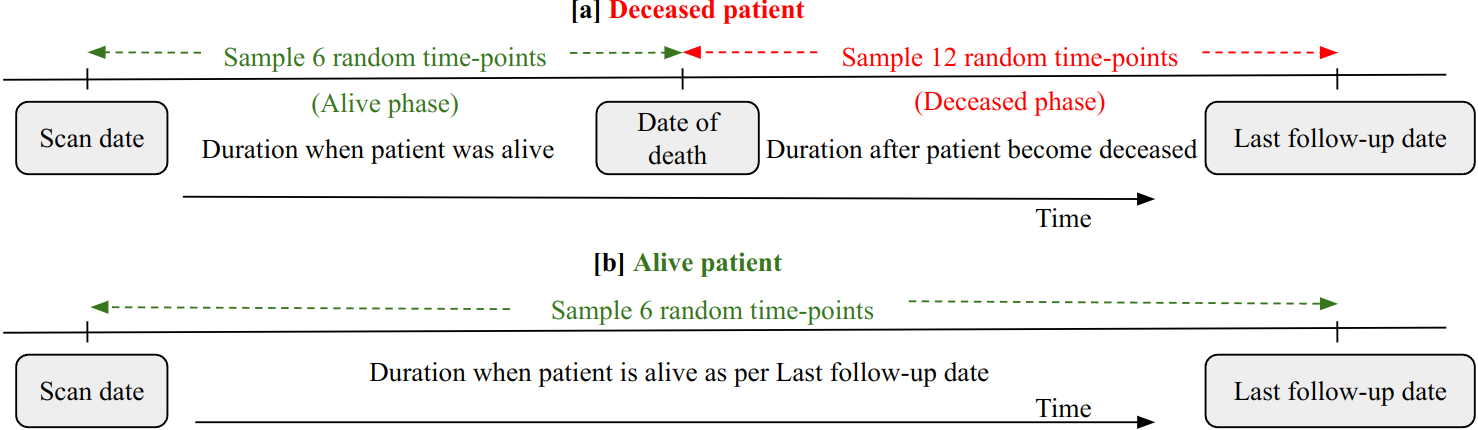} % Replace "image_filename" with the actual file name of your image
  \caption{Data sampling strategy used during training of the proposed framework. [a] For deceased patients, random time-points were sampled from two phases: the alive phase (time between scan date and the date of death) and the deceased phase (time between the date of death and LFD). [b] For alive patients, random time-points were sampled between the scan date and LFD.}
  %\label{fig:labelname}
  
\end{figure}

For patients who remained alive throughout the observation window (see Figure 2[b]), 6 random time-points per epoch were sampled from the observation window, with the corresponding OS statuses labeled as "alive." 
For these patients, time-points were sampled only up to the LFD, and no samples were generated beyond this point, as survival status is unknown beyond follow-up.
For deceased patients (see Figure 2[a]), the observation window was divided into two phases: "alive phase" (from scan date to date of death) and "deceased phase" (from date of death to LFD). 
Since the exact time of death is known, time-points across the full observation window were used for these patients.
During training, 6 random time-points were sampled from the alive phase, and 12 random time-points were sampled from the deceased phase, with the corresponding OS statuses and the network weights updated accordingly during training. This resulted in approximately 2500–2700 time-points per epoch for both classes, ensuring a balanced representation of alive and deceased phase during training. 
A detailed description of the time-specific sampling strategy is shown in Figure 2, and key terminologies are defined in \ref{appendix:key_term_data_sampling}. 

Random sampling of time-points was chosen to introduce variability during training, which improved the model’s generalizability. The network was trained to classify the OS status for each sampled time-point, enabling it to predict OS probabilities as a function of time. By sampling different time-points in each epoch, the risk of overfitting to specific time intervals was reduced. 
This formulation enables time-conditioned survival prediction, where identical imaging inputs are evaluated at different time horizons, and therefore does not introduce redundancy.
We also experimented with uniform sampling, but found that random sampling led to better generalization. 
The choice of 6 and 12 time-points was empirically determined based on preliminary experiments, providing a balance between temporal resolution, class imbalance handling, and computational efficiency.
During validation/testing, the model was evaluated at fixed time-points at 30-day intervals from the scan date till the LFD, and the evaluation metrics were calculated based on the predicted OS probabilities. This approach ensured a consistent evaluation strategy across patients. The loss function used during training was a combination of focal loss and survival consistency loss, as described in \ref{appendix:loss_func}.

\subsection{\textbf{Experimentation}}

To evaluate the impact of temporal data, a baseline approach using only the ResNet-50 backbone was implemented to classify OS status as deceased or alive at pre-specified intervals. The baseline approach consisted of 10 separate models trained at pre-specified intervals ranging from 0.5 years to 5 years, with 6-month intervals between them. In contrast, the proposed method used a single model, trained using the same data but incorporated additional time-points as specified in Section 3.3. A detailed comparison between the baseline and proposed methods is provided in Table \ref{tab:comp_train_bl_p1_arch}. Both models (baseline and proposed) were trained for 100 epochs using data augmentation, optimized using the Adam optimizer with an initial learning rate of 1e-4, a weight decay of 1e-5, and a learning rate scheduler that reduces the learning rate by a factor of 5 after every 5 epochs if the validation loss does not decrease. Both methods were developed and validated using 
patient-level stratified
5-fold cross-validation (n=556), and tested on a held-out test set (n=292). The results were then compared at pre-specified intervals. 

\begin{table}[!htb]
\centering
\caption{Comparison criteria for each network between baseline and proposed methods for OS prediction across various criteria. In both the methods ResNet-50 was used as the backbone architecture. Here "n" corresponds to number of networks used (In our case, n = 10).}
\label{tab:comp_train_bl_p1_arch}
%\tiny
\scriptsize
\begin{tabular}{lcc}
\hline
\multirow{2}{*}{Comparison criteria (for each network)} & \multicolumn{2}{c}{Deep regression (ResNet-50)}  
\\
\cline{2-3}
                        & baseline & proposed\\
\hline
{[}1{]} Number of networks      & n                   & 1   \\
\hline
{[}2{]} Trainable parameters for each network      & 24,680,769                  & 33,077,569                   \\
\hline
{[}3{]} Model size (MB) for each network      & 296                  & 397   \\
\hline
{[}4{]} Training time per epoch (seconds)      &     10              & 225   \\
\hline
{[}5{]} Validation time per epoch (seconds)      &   2                & 126   \\
\hline
{[}6{]} RAM usage      & Low                  & High   \\
\hline
{[}7{]} Overall performance with current hardware      & Poor                   & Good   \\
\hline
{[}8{]} Overfitting      & More prone                   & Less prone   \\
\hline
\end{tabular}
\end{table}

To further contextualize the performance of the proposed method, we benchmarked it against two established survival analysis models from the Pycox library \cite{kvamme2019time}: DeepHit (a discrete-time model for single events) \cite{lee2018deephit} and DeepSurv (a CPH-based deep learning extension) \cite{katzman2018deepsurv}. Both methods were trained using two different approaches. The first approach employed conventional clinical features
(clin) such as age, sex, TN staging
along with explicit imaging-derived phenotypes (IDP) such as TMTV, Dmax, 
Lesion count,
as input, referred to as DeepHit (clin + IDP) and DeepSurv (clin + IDP). The second approach utilized image-derived embeddings generated using a CNN backbone, such as ResNet-50, and is referred to as DeepHit (imaging) and DeepSurv (imaging). Both methods were trained on the same dataset (n=556 for training/validation, n=292 for testing) under identical cross-validation splits. The implementation details for the DeepHit and DeepSurv methods are provided in \ref{appendix:clin_IDP}.

The performance of all methods (baseline, proposed, DeepHit, and DeepSurv) was evaluated at pre-specified time intervals. The reported results, included the area under the curve (AUC), C-index, and accuracy. These values represented the mean performance obtained by applying five cross-validated models to the test set, which helped reduce variability in performance metrics. Ablation experiments were conducted only for the proposed method by removing the input image channels, to assess the effectiveness of tissue-specific FDG-PET/CT (see Table \ref{tab:tissue_wise}). Saliency analysis using Grad-CAM \cite{selvaraju2017grad} was performed to visualize important image regions contributing to the network's prediction. Kaplan-Meier curves for the test set and OS probabilities as a function of time for each patient were generated to perform a qualitative assessment.

\subsection{\textbf{Risk stratification}}

Risk stratification is a method of categorizing patients into distinct prognostic groups (risk levels) based on their predicted likelihood of specific outcomes, in this case OS. Using the proposed method, patient-specific OS probability curves as a function of time were generated, as shown in Figure 3. To quantify risk, we introduce a metric termed as the Area Under the Survival Probability Curve (AUSPC), computed by integrating the predicted survival probability over time;
which is conceptually related to the restricted mean survival time (RMST) used in survival analysis \cite{royston2013restricted}.
In this study, the AUSPC was calculated over a time horizon of 5 years using the trapezoidal numerical integration method applied to the predicted survival probability curve.
This metric served as an inverse risk indicator, i.e., higher AUSPC values corresponded to better prognosis (lower risk), while lower values indicated poorer prognosis (higher risk), as illustrated in Figure 3 ([a] low-risk survivor vs [b] high-risk case). For population-level analysis, AUSPC was computed for all patients within the test set, and k-means clustering (k=2) was applied, in order to partition the cohort into two distinct risk categories.
This unsupervised clustering was performed directly on the AUSPC values and automatically separated patients into high-risk and low-risk groups without using a predefined threshold (e.g., median OS).

\begin{figure}[!htb]
  \centering
  \includegraphics[width=0.95\textwidth]{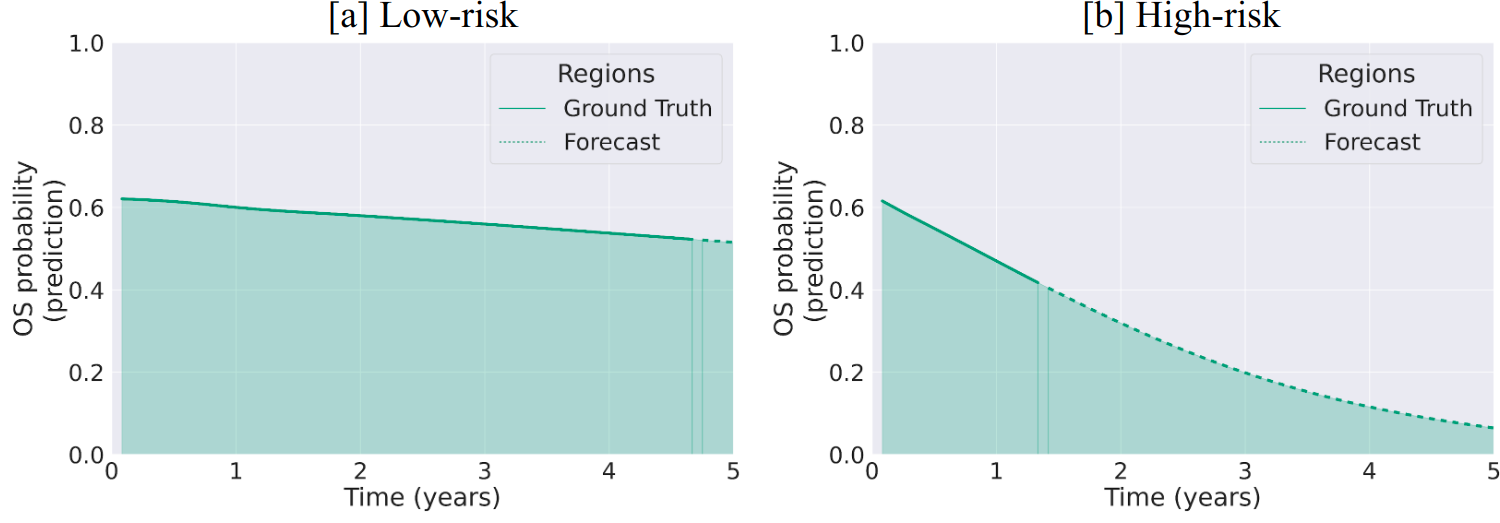}
  \caption{\small Illustration of OS probability over time for patients in different risk categories: (a) Low-risk patient: OS probability remains high and stable (high AUSPC). (b) High-risk patient: OS probability drops quickly (low AUSPC).}
\end{figure}

Risk stratification was additionally performed within each T-stage category (T1, T2, T3, T4) using the cross-validation cohort, because it was larger than the test set and it provided sufficient sample sizes for robust subgroup analysis. Application to the test set was not feasible due to limited sample sizes, where further stratification would have resulted in groups too small for meaningful statistical analysis. Finally, Kaplan-Meier curves were generated, with patients stratified into different risk categories (high risk vs low risk) based on model predictions. 
The subgroup sizes used for Kaplan-Meier analysis were as follows: T1 (high-risk = 15, low-risk = 165), T2 (high-risk = 26, low-risk = 89), T3 (high-risk = 22, low-risk = 51), and T4 (high-risk = 37, low-risk = 57).
Differences between survival curves were evaluated using the log-rank test, with a $p-value < 0.05$ considered statistically significant.

\section{Results}

Table \ref{tab:bl_vs_pro} summarizes the results of clinical and imaging-based DeepHit/DeepSurv variants along with the baseline and proposed methods in predicting OS at pre-specified intervals. The baseline and proposed methods used a ResNet-50 backbone. The proposed method achieved superior performance (0.746) compared to both baseline (0.703) and other benchmark approaches such as DeepHit (clin + IDP) (0.693), DeepSurv (clin + IDP) (0.717), DeepHit (imaging) (0.683), and DeepSurv (imaging) (0.689). 

In addition, we evaluated the contribution of different input modalities. While models using clinical and IDP features alone achieved strong performance, the proposed imaging-only model showed competitive results. Importantly, an ensemble of the imaging and clinical + IDP models achieved the best performance across all time points (0.788), outperforming both individual models. This suggests that imaging features provide complementary information that enhances survival prediction when integrated with clinical and IDP features.

\begin{table}[!htb]
\centering
\caption{\small Comparison between the baseline and proposed methods in predicting OS status at pre-specified time intervals from the test set.}

\label{tab:bl_vs_pro}
\scriptsize % Change the font size to \scriptsize
\begin{tabular}{p{4.9cm}ccccccccccc}
\hline
\multirow{2}{*}{Experimentation} & \multicolumn{11}{c}{AUC at pre-specified time intervals (in years)}\\
\cline{2-12}
& .5 & 1 & 1.5 & 2 & 2.5 & 3  & 3.5  & 4 & 4.5 & 5 & Mean \\
\hline
(1) CPH (clin + IDP) & .583 & .580 & .647 & .648 & .662 & .664 & .657 & .683 & .677 & .693 & .649 \\
\hline
(2) DeepHit (clin + IDP) & .750 & .706 & .727 & .740 & .736 & .694 & .662 & .669 & .641 & .604 & .693 \\
\hline
(3) DeepSurv (clin + IDP) & \textbf{.773} & .719 & .729 & .734 & .735 & .719 & .711 & .718 & .719 & .714 & .717 \\
\hline
(4) DeepHit (imaging) & .751 & .705 & .701 & .677 & .675 & .662 & .662 & .666 & .666 & .660 & .683 \\
\hline
(5) DeepSurv (imaging) & .724 & .684 & .698 & .698 & .693 & .684 & .676 & .679 & .678 & .678 & .689 \\
\hline
(6) baseline (imaging) & .630 & .680 & .703 & .732 & .698 & .715 & .696 & .713 & .718 & .750 & .703 \\
\hline
(7) proposed (imaging) & .694 & .710 & .761 & .760 & .754 & .747 & .742 & .762 & .756 & .772 & .746 \\
\hline
(8) proposed (clin + IDP) & .742 & .733 & .753 & .762 & .782 & .782 & .764 & .796 & .792 & .807 & .771 \\
\hline
(9) proposed ensemble & .760 & \textbf{.750} & \textbf{.782} & \textbf{.789} & \textbf{.795} & \textbf{.794} & \textbf{.780} & \textbf{.808} & \textbf{.802} & \textbf{.817} & \textbf{.788} \\
\hline
\end{tabular}
\end{table}

%\newpage

Table \ref{tab:tissue_wise} presents the results of the ablation experiments using the proposed method (with a ResNet-50 backbone) on the test set, using various combinations of tissue-wise FDG-PET/CT and tumor projections. Inclusion of the tumor projection channel achieved the highest OS prediction performance, with an overall AUC of 0.826, a C-Index of 0.733, and an accuracy of 0.738. This is consistent with previous studies, showing a strong correlation between TMTV and clinical outcomes such as OS \cite{girum202218f} \cite{mikhaeel2022proposed}. 

\begin{table}[!htb]
\centering
\caption{\small Results of ablation experiments using the proposed method on the test set, illustrating the impact of various tissue-wise FDG-PET/CT input channels on OS prediction.}

\label{tab:tissue_wise}
\scriptsize % Change the font size to \scriptsize
\begin{tabular}{p{2.9cm}cccccp{.6cm}p{1cm}p{.6cm}p{.6cm}p{.6cm}}
\hline
\multirow{2}{*}{Experimentation} & \multicolumn{5}{c}{Inputs ($CT^{\text{AIP}}$/$PET^{\text{MIP}}$/$SEG$)} & \multirow{2}{*}{AUC} & \multirow{2}{*}{C-Index} & \multirow{2}{*}{Accuracy} \\
\cline{2-6}
& org & bone & lean & adipose & air &   &   &   \\
\hline
1 org PET & $\times$/$\checkmark$/$\times$ & $\times$/$\times$/- & $\times$/$\times$/- & $\times$/$\times$/- & $\times$/$\times$/- & .759 & .657 & .670  \\
\hline
2 org CT & $\checkmark$/$\times$/$\times$ & $\times$/$\times$/- & $\times$/$\times$/- & $\times$/$\times$/- & $\times$/$\times$/- & .758 & .613 & .683  \\
\hline
3 all PET & $\times$/$\checkmark$/$\times$ & $\times$/$\checkmark$/- & $\times$/$\checkmark$/- & $\times$/$\checkmark$/- & $\times$/$\checkmark$/- & .770 & .635 & .677  \\
\hline
4 all CT & $\checkmark$/$\times$/$\times$ & $\checkmark$/$\times$/- & $\checkmark$/$\times$/- & $\checkmark$/$\times$/- & $\checkmark$/$\times$/- & .752 & .575 & .677  \\
\hline
5 all PET/CT  & $\checkmark$/$\checkmark$/$\times$ & $\checkmark$/$\checkmark$/- & $\checkmark$/$\checkmark$/- & $\checkmark$/$\checkmark$/- & $\checkmark$/$\checkmark$/- & .796 & .643 & .705  \\
\hline
6 all PET/CT-SEG  & $\checkmark$/$\checkmark$/$\checkmark$ & $\checkmark$/$\checkmark$/- & $\checkmark$/$\checkmark$/- & $\checkmark$/$\checkmark$/- & $\checkmark$/$\checkmark$/- & \textbf{.826} & \textbf{.733} & \textbf{.738}  \\
\hline
\end{tabular}
\end{table}

Figure 4[a] illustrates the predicted OS probabilities as a function of time for a patient with a GT OS of 0.53 years and a predicted OS of 0.50 years. Figure D.8[a]-[d] in \ref{appendix:OS_examples} show examples of OS prediction plots for deceased patients (who died during the follow-up period). Such plots could aid in future forecasting of OS probabilities as a function of time, based on medical images. Figure D.8[e]-[h] in \ref{appendix:OS_examples} show examples of future forecast of OS probabilities for patients who remained alive, as per LFD. The forecasted OS probabilities are represented as dotted lines: green for the alive phase and red for the deceased phase. These plots visualize the network's prediction of OS probabilities and can highlight trends or discrepancies, where deviations may indicate variations in disease progression or treatment response among patients. It could potentially enhance clinical decision-making by providing individualized survival predictions, with an aim of improving patient outcomes. The potential of forecasting OS probabilities over time could allow for better monitoring of disease progression and early adjustments to treatment plans. 

\begin{figure}[!htb]
  \centering
  \includegraphics[width=0.90\textwidth]{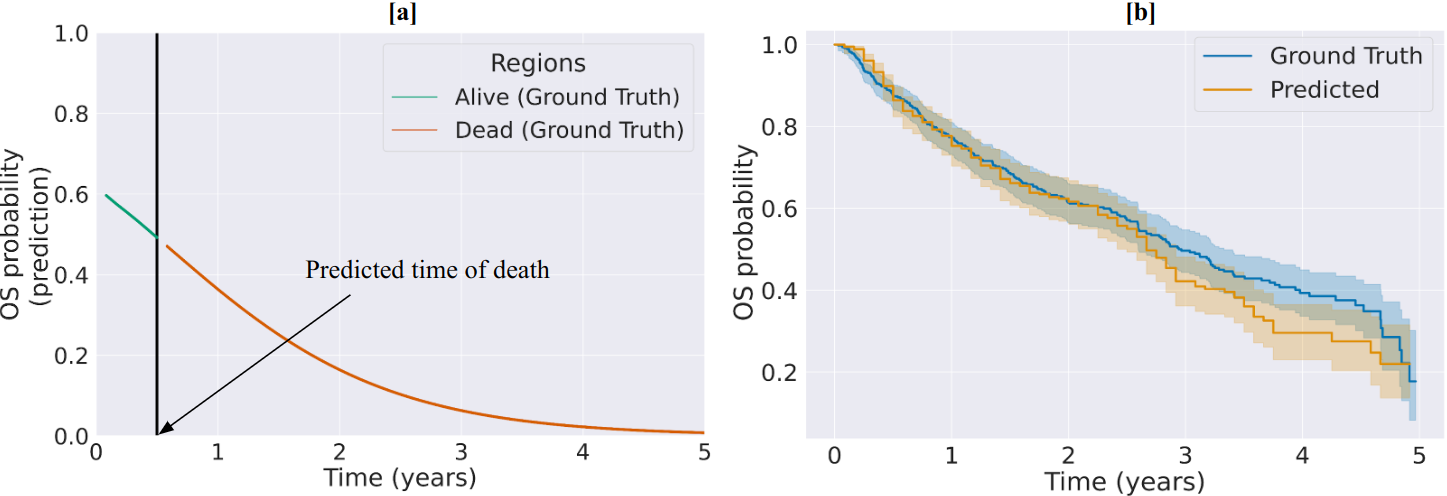}
  \caption{\small (a) Predicted OS probabilities as a function of time for a patient where the green line represents the alive phase (GT), the red segment represents the deceased phase (GT), and the intersection of the black line with the x-axis indicates the predicted time of death. (b) Kaplan-Meier curve illustrating GT vs predicted OS probabilities on the test set.}
\end{figure}

Figure 4[b] illustrates the Kaplan-Meier curve, comparing the GT and predicted survival times on the test set. The predicted curve closely resembles the GT curve with a slight underestimation of OS probabilities at later time-points, indicating the proposed method's effectiveness with a minor bias in the long-term survival estimation. Figure E.9 in \ref{appendix:KM_group_wise} shows the comparison of Kaplan-Meier curves for [a] low vs high TMTV and [b] female vs male in NSCLC. It demonstrates that patients with high TMTV have poorer survival outcomes compared to those with low TMTV, which is in line with previous studies \cite{seban2020baseline}. Additionally, males were shown to have at a higher risk compared to females.

Figure 5 shows Kaplan-Meir survival curve, illustrating the stratification of NSCLC patients in the test set into two distinct risk categories using the proposed method. The proposed method was able to group the patients into low-risk (n = 252) and high-risk categories (n = 40), with the low-risk group exhibiting significantly longer survival compared to the high-risk group ($p < 0.05$). This highlights the robust prognostic capability of our AI-driven approach, offering clinically relevant risk categories. 

\begin{figure}[!htb]
  \centering
  \includegraphics[width=0.60\textwidth]{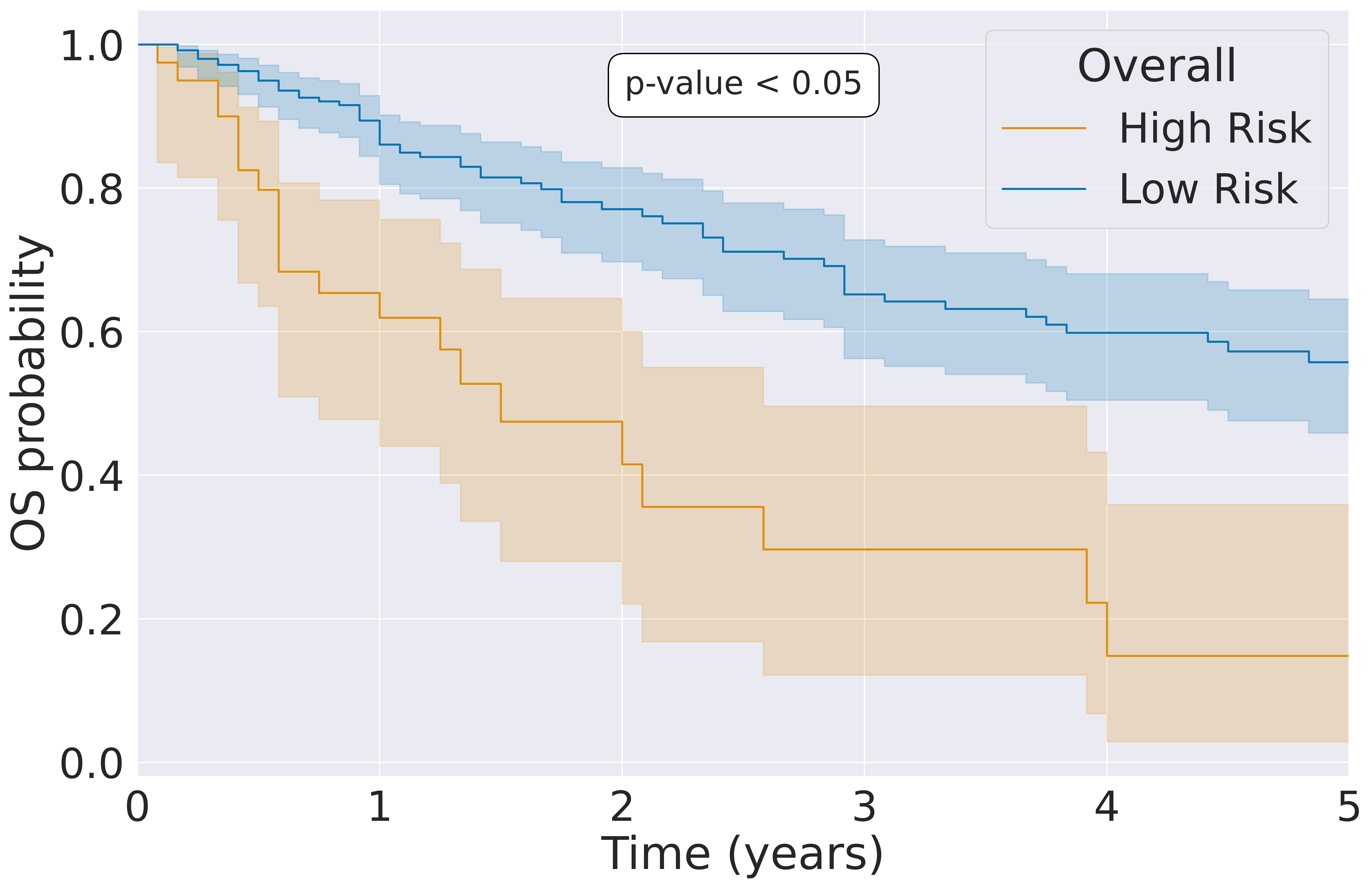}
  \caption{\small Kaplan–Meier curve illustrating risk stratification of NSCLC patients in the test set, based on model-predicted survival probabilities, dividing them into high-risk (n=40) and low-risk (n=252) groups.}
\end{figure}

Figure 6 [a]–[d] shows Kaplan-Meir survival curves, illustrating the risk stratification of NSCLC patients within each T-staging (T1–T4) in the cross-validation set. Significant differences in survival were observed between high- and low-risk groups for T2, T3, and T4 ($p < 0.05$), while the difference for T1 was not statistically significant ($p > 0.05$).

\begin{figure}[!htb]
  \centering
  \includegraphics[width=0.99\textwidth]{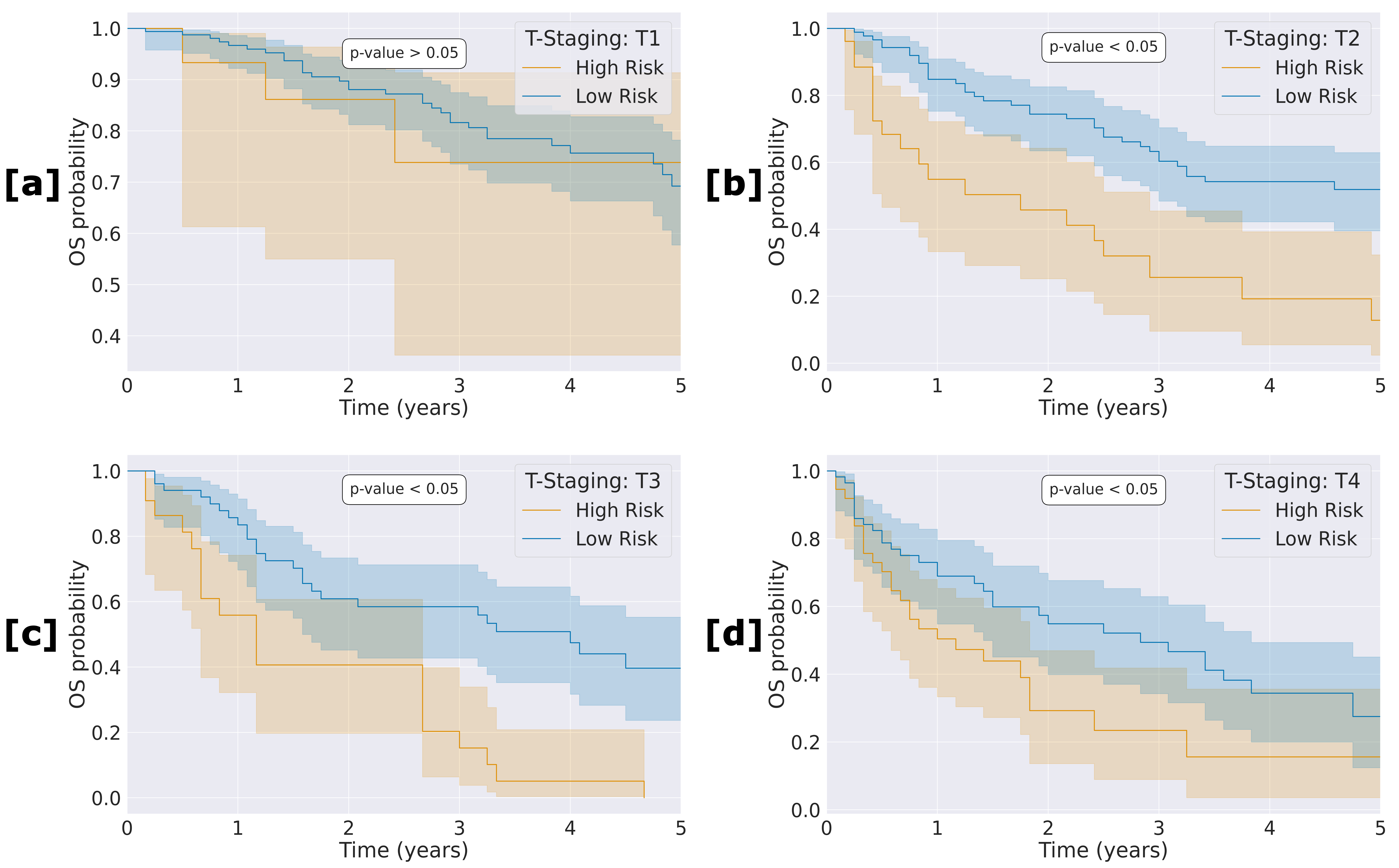}
  \caption{\small Kaplan–Meier curves illustrating risk stratification of NSCLC patients within each T-stage in the cross-validation set using the proposed method: (a) T1 (high-risk = 15; low-risk = 165), (b) T2 (high-risk = 26; low-risk = 89), (c) T3 (high-risk = 22; low-risk = 51), and (d) T4 (high-risk = 37; low-risk = 57), based on model-predicted survival probabilities.}
\end{figure}

%\newpage

Figure 7 presents the  saliency maps for patients with varying OS outcomes. It shows that the network predominantly focuses on the tumor regions when predicting OS. This pattern was consistently observed throughout the U-CAN cohort, with a few exceptions.

\begin{figure}[!htb]
  \centering
  \includegraphics[width=1.00\textwidth]{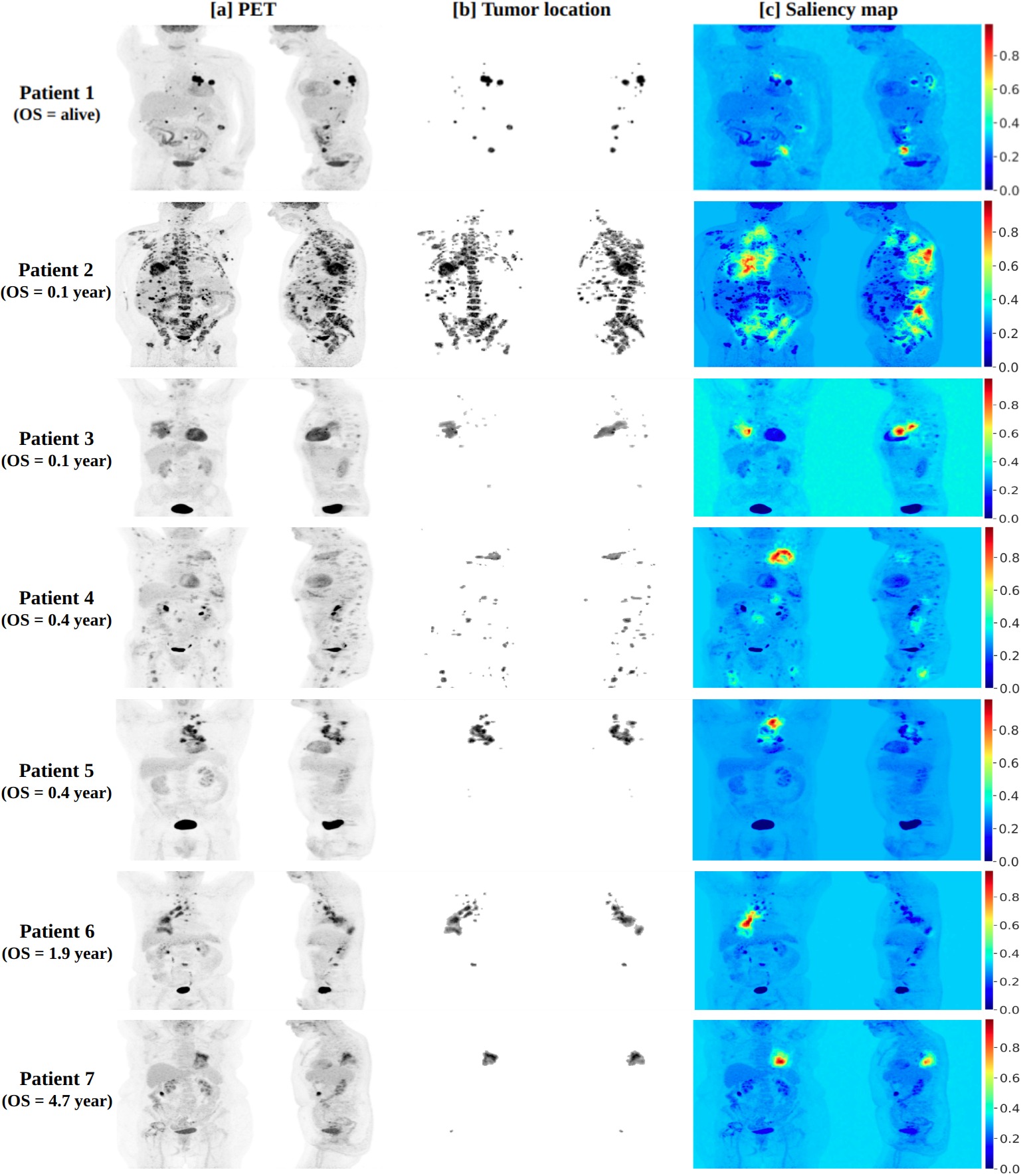}
  \caption{\small Overview of saliency analysis from the proposed method: [a] PET MIP, [b] Tumor location, [c] Heatmap highlighting regions influencing the neural network's decision.}
\end{figure}

\section{Discussion}

In this study, we have developed a deep learning based time-driven survival analysis method for predicting OS as a function of time, followed by risk stratification. The use of 2D projections was motivated by the need for computational efficiency and to mitigate the risk of overfitting due to the limited data and high-dimensional 3D PET/CT inputs.

Our results demonstrated that training separate networks for different time intervals (as done in the baseline approach) is not only redundant and costly but also failed to fully utilize the dataset's potential. Instead, our proposed method employed a single network, trained on multiple time-points, to effectively model the OS probability as a function of time with improved generalization on unseen time-points. The proposed method also better captured temporal dependencies in survival prediction compared to either discrete-time (DeepHit) or continuous-time (DeepSurv) alternatives.
At shorter time horizons, slightly lower performance compared to DeepSurv may be attributed to the limited number of early events, where risk-ranking approaches such as DeepSurv can achieve strong discrimination.
The performance advantage was particularly notable at longer time horizons (1.5-5 years), where the proposed method maintained consistently higher AUC values (0.747-0.772) compared to other individual models (see Table \ref{tab:bl_vs_pro}). Notably, the ensemble of imaging and clinical + IDP models achieved the best overall performance across all time points.

While clinical and IDP features achieved strong performance, the ensemble results indicate that imaging-derived features provide complementary prognostic information not fully captured by these variables. The combined model consistently outperformed both individual models, suggesting that neither representation alone is sufficient to fully characterize patient risk.
In this context, tumor segmentation plays an important role, as segmentation-derived channels provide tumor-specific information; therefore, the accuracy of tumor delineation can influence prediction performance.

The baseline method predicted OS at pre-specified intervals and relied heavily on right censoring, sometimes excluding patients lost to follow-up. It also treated patients who did not survive beyond the fixed-time mark equivalently, regardless of whether they pass away shortly or much later. Such binary classification failed to capture the gradual change in OS probability over time, which could lead to information loss, reduced sample size, and a limited ability to model the complexity of the task. In contrast, the proposed method enabled flexible OS predictions at arbitrarily chosen time-points, reducing the need for right censoring and effectively parameterizing the predictions based on time. By incorporating temporal information, the framework can predict OS probabilities at any given time, rather than being constrained to pre-specified intervals. 

The integration of temporal data ensures that the model can generalize across diverse survival patterns and handle censored cases more effectively. By training on randomly sampled time-points, the proposed method was able to provide a robust and comprehensive representation of patient survival patterns. The choice of 6 (from alive phase) and 12 (from deceased phase) time-points was made to balance the representation of both phases. Since the number of deceased patients is smaller than alive patients, it led to a class imbalance between the deceased and alive time-points. To address this, oversampling from the deceased phase was implemented, ensuring that the model was adequately exposed to the transition from alive to deceased status. We experimented with other numbers of sampling points but settled on 6 and 12 as they provided a realistic balance between computational feasibility and model performance.

The results presented in Figures 5 and 6 indicated that the proposed method has the potential to effectively stratify NSCLC patients into clinically relevant risk categories (high vs low risk) across the entire cohort. Such differentiation holds great potential to support personalized treatment planning. In particular, when stratification was performed within each T-staging, it was found to be statistically significant for stages T2 through T4, but not for T1. This observation aligns with clinical intuition, as early-stage tumors (T1) typically have uniformly favorable outcomes, whereas more advanced stages (T2–T4) exhibit greater prognostic heterogeneity, further influenced by variability in treatment approaches \cite{flury2022heterogeneity}. In this study, our focus was on risk stratification based on T-staging, which is a key component for cancer prognostics. Although N and M staging (representing nodal involvement and metastasis) are also critical prognostic factors, they were not included in the current analysis. Future work will extend the stratification framework to incorporate N and M staging, enabling a more comprehensive assessment of patient risk profiles.

Saliency analysis helped visualize the most influential image regions driving model predictions, improving transparency and helping identify key factors in patient outcomes. The saliency maps as shown in Figure 7 were derived from the convolutional layers of the ResNet-50 backbone. These heatmaps were generated before incorporating temporal data, which was processed separately through a FNN and combined with the image embeddings. While this process refines OS predictions based on time-dependent context, it does not directly influence the spatial regions identified in the heatmaps. It only modulates the image embeddings to account for the time-dependent nature of OS probabilities. 
In our experiments, heatmaps generated with and without temporal inputs exhibited largely similar spatial patterns based on qualitative assessment. This observation should not be interpreted as evidence of strict independence, and a more rigorous evaluation of the influence of temporal modeling on saliency maps remains an area for future work.

One limitation of the proposed method is its reliance on training across multiple time-points to achieve adequate generalization. As a result, the training time for the proposed method is higher than that of the baseline method, as shown in Table \ref{tab:comp_train_bl_p1_arch}.
%Table F.4 of Appendix F. 
While the baseline method provided a simple yet efficient framework, it was prone to overfitting. The proposed method on the other hand was more complex, required greater computational resources but delivered improved performance with better generalization. As a result, the current implementation of the proposed method incurred higher computational costs. Therefore, future work will focus on enhancing the proposed method, specifically by incorporating the temporal aspect into the model while reducing computational demands.
It will also explore the use of more advanced deep learning architectures and hybrid frameworks for improved feature extraction and multimodal integration, to further enhance survival prediction performance.

Additionally, the study is based on a single-institution dataset without external validation, which may limit generalizability to other clinical settings. Future work will focus on validation using multi-institutional datasets.
Some clinical variables also contained missing values, which may affect the completeness of tabular feature representation.

\section{Conclusion}

We introduced an automated framework that incorporated image embeddings from tissue-wise FDG-PET/CT projections with temporal data to predict OS as a function of time in NSCLC patients. The proposed method outperformed the baseline method as well as well-established survival models with higher AUC at pre-specified intervals (1.5-5 years). 
Furthermore, an ensemble of imaging and clinical + IDP models achieved the best overall performance, highlighting the complementary nature of these feature representations.
Risk stratification using model-derived survival probabilities identified distinct prognostic groups with significantly different outcomes. Saliency analysis highlighted tumor regions as key predictors of OS.

\section{\textbf{Declarations}}

\subsection{\textbf{Ethics approval and consent to participate}}

Ethical approval to conduct retrospective image analysis on the U-CAN dataset was obtained from the Swedish Ethical Review Authority (Dnr 2023-02312-02). The study was conducted in accordance with relevant guidelines and regulations, including the Declaration of Helsinki.

\subsection{\textbf{Acknowledgement}}

We would like to thank Alexander Korenyushkin for sharing insights on image interpretation and 3D Slicer. We would also like to thank Nikolaos Kalampokas and Aditya Harichandar for their contributions to the development of the U-CAN FDG-PET/CT preprocessing pipeline and the selection of the U-CAN data. 

\subsection{\textbf{Availability of data and code}}

UCAN data is available upon approval of dedicated application to the UCAN scientific evaluation committee. The code is available at: \url{https://github.com/ashishch-git/OS_prediction}.

\subsection{\textbf{Competing interests}}

The authors disclose no competing interests.

\subsection{\textbf{Funding}}

This study was supported by the Swedish Cancer Society (23 3123 Pj 01 H), Analytic Imaging Diagnostics Arena (AIDA), Lions cancer fund and Makarna Eriksson foundation.

\subsection{\textbf{Author contributions}}

Conceptualization: ST, AC, JO and JK; Method development: ST, AC, EL, JO and JK; Image analysis: ST, AC, EL, JO, TS, VSR and JK. Interpretation of results: ST, AC, EL, JO, TS and JK; Data Annotation: VSR; Funding acquisition: HA and JK; Supervision: EL, HA, JK; Manuscript preparation: ST and AC; Manuscript revision, reading, and approval of the final version: ST, AC, EL, JO, TS, VSR, HA, JK.

%\bibliographystyle{vancouver-fullpages.bst}

%\bibliographystyle{plainnat} % This style will give the correct formatting
%\bibliography{references} 

\bibliographystyle{elsarticle-num} 
\bibliography{references}

\newpage
\appendix

\section{Data description} \label{appendix:data_desc}

\begin{table}[!htb]
\centering
%\scriptsize % Change the font size to \scriptsize
\small
\caption{Data description for NSCLC patients in the U-CAN cohort.}
\begin{tabular}{|p{4.5cm}|p{3.5cm}|p{3.5cm}|}
\hline
& \textbf{Cross-val (missing)} & \textbf{Test (missing)} \\
\hline
\textbf{Total} & \textbf{556} & \textbf{292} \\
\hline
\multicolumn{3}{|l|}{\textbf{Demographics}} \\
\hline
\textbf{Age (years)} & 70.31 $\pm$ 8.62 & 69.81 $\pm$ 10.31 \\
\hline
\textbf{Sex} & \textbf{556} & \textbf{292} \\
Male & 239 & 127 \\
Female & 317 & 165 \\
\hline
\multicolumn{3}{|l|}{\textbf{Clinical Data}} \\
\hline
\textbf{T-Staging} & \textbf{462 (17\%)}  & \textbf{178 (39\%)} \\
I & 180 & 83 \\
II & 115 & 43 \\
III & 73 & 22 \\
IV & 94 & 30 \\
\hline
\textbf{N-Staging} & \textbf{455 (18\%)} & \textbf{176 (40\%)} \\
0 & 271 & 129 \\
I & 39 & 10 \\
II & 85 & 20 \\
III & 60 & 17 \\
\hline
\textbf{M-Staging} & \textbf{385 (31\%)} & \textbf{119 (59\%)} \\
0 & 278 & 93 \\
I & 107 & 26 \\
\hline
\textbf{Treatment Received} & \textbf{278 (50\%)} & \textbf{78 (73\%)} \\
CHEMOTHERAPY & 186 & 47 \\
RADIOTHERAPY & 57 & 16 \\
IMMUNOTHERAPY & 17 & 11 \\
OTHER & 18 & 4 \\
\hline
% \multicolumn{3}{|l|}{\textbf{Imaging Data}} \\
% \hline
% \textbf{PET-CT Parameters} & & \\
% voxel size & [2.04, 2.04, 3.0] &  \\
% image size & (512, 512, 480) & \\
% \hline
%Tumor Segmentation & N/A & X\% \\
%SUV Metrics & 0 -- 40 & X\% \\
\textbf{Image-derived Features} &  & \\
TMTV (ml) & 54.45 $\pm$ 98.85 & 45.36 $\pm$ 80.62 \\
Dmax (mm) & 95.94 $\pm$ 77.08 & 55.60 $\pm$ 78.32 \\
\hline
\multicolumn{3}{|l|}{\textbf{Outcome}} \\
\hline
\textbf{Survival Status} & \textbf{Deceased / Alive} & \textbf{Deceased / Alive} \\
1 year & 100 / 405 & 43 / 173 \\
2 year & 159 / 299 & 59 / 85 \\
3 year & 195 / 229 & 73 / 70 \\
4 year & 218 / 145 & 80 / 55 \\
5 year & 228 / 92 & 84 / 34 \\
\hline
\end{tabular}
\end{table}

\newpage

\section{Key terminologies used during data sampling strategy} \label{appendix:key_term_data_sampling}

The data sampling strategy utilized during the training of the proposed framework is illustrated in Figure 2. Key terminologies used for data sampling strategy are defined below:

\begin{itemize}
    \item \textbf{Scan date:} The date when the FDG-PET/CT data was acquired for a patient.
    
    \item \textbf{Last follow-up date (LFD):} The most recent date on which the survival status of all patients was last verified.

    \item \textbf{Observation window:} The period between the scan date and the LFD.
    
    \item \textbf{Deceased patient:} A patient who died during the observation window. Patient data included both the alive and deceased phases.
    
    \item \textbf{Alive patient:} A patient who remained alive throughout the observation window, as per the LFD.
    
    \item \textbf{Date of death:} The date when a patient was confirmed to have died. For deceased patients, this date marks the boundary between the alive and deceased phases.

    \item \textbf{Alive phase:} The period from the scan date until the date of death (for deceased patients) or LFD (for alive patients). Random time points sampled from this phase were labeled as "alive."
    
    \item \textbf{Deceased phase:} For deceased patients, this is the period from the date of death to the LFD. Random time points sampled from this phase were labeled as "deceased."
\end{itemize}

\newpage
\section{Loss function} \label{appendix:loss_func}

To handle the class imbalance problem in the binary classification task, Focal loss \cite{ross2017focal} was used. It is defined as:

\begin{equation} \label{eq:FocalLoss}
    \text{Focal loss}(p_t) = -\alpha_t (1 - p_t)^\gamma \log(p_t)
    \tag{5}
\end{equation}

where:
\begin{itemize}
    \item $p_t$ is the model's estimated probability for the true class label.
    \item $\alpha_t$ is a balancing factor for the class t, which adjusts the weight of positive and negative examples. 
    \item $\gamma$ is a focusing parameter that controls the rate at which easy examples are down-weighted. When $\gamma = 0$, focal loss is the same as cross-entropy loss. As $\gamma$ increases, the focus shifts from easy examples to hard-to-classify ones. \\
\end{itemize}

In case of the proposed method, to prevent the increase in OS probability over time, Survival consistency loss (SCL) was introduced. This helped ensuring that the probability of OS remains stable and consistent throughout the training process. It is defined as:
\begin{equation}
\mathtt{SCL} = max(0, p(t_2)-p(t_1))
\tag{6}
\end{equation}
where $p(t_1)$ denotes OS probability at time-point $t_1$,
$p(t_2)$ denotes OS probability at time-point $t_2$, and
$t_2$ \textgreater $t_1$.\\

The final loss was the combination of Focal loss and SCL given by:
\begin{equation}
\text{Final loss} = \text{Focal loss} + \lambda * \text{SCL}. \\
\tag{7}
\end{equation}
where $\lambda$ is a weighting factor ($\lambda$ = 1 in our case).\\

In case of baseline method, only focal loss was used as the loss function during optimization. In case of proposed method, a combination of focal loss and survival consistency loss was used during optimization.

\newpage
\section{Overall survival probabilities as a function of time} \label{appendix:OS_examples}

\begin{figure}[!htb]
  \centering
  \includegraphics[width=0.73\textwidth]{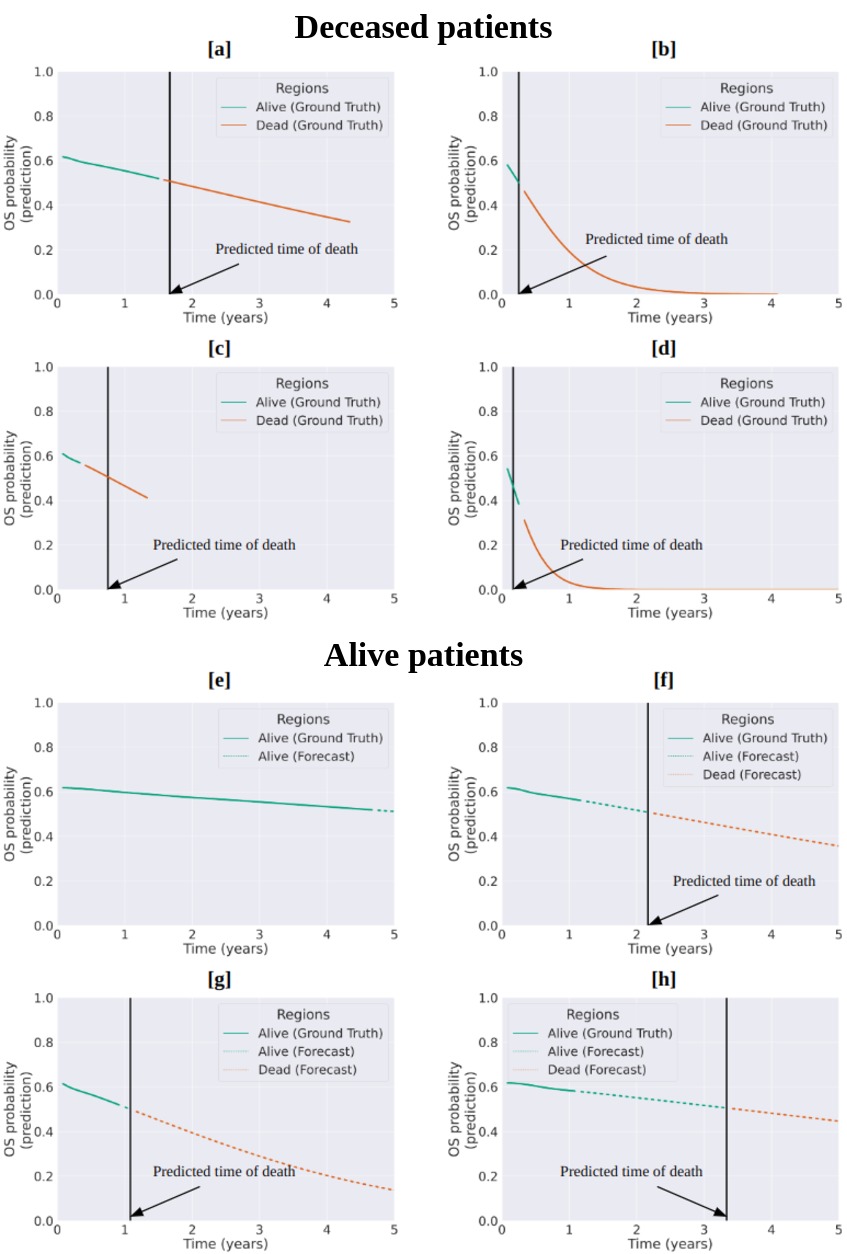}
  \caption{\small Overview of predicted OS probabilities as a function of time on the test set for patients who were deceased ([a]-[d]) and patients who remained alive ([e]-[h]) as per LFD. Solid green line represents the patient in ground truth (GT) alive phase. Solid red lines represents the patient in GT deceased phase. Dotted green lines indicate the future forecast for patients in the alive phase. Dotted red lines indicate the future forecast for patients in the deceased phase. The point where the solid black line intersects the x-axis indicates the predicted time of death.}
  \label{fig:appendix_figure}
\end{figure}

\newpage
\section{Kaplan-Meier curves with respect to patient characteristics} \label{appendix:KM_group_wise}

\begin{figure}[!htb]
  \centering
  \includegraphics[width=1.0\textwidth]{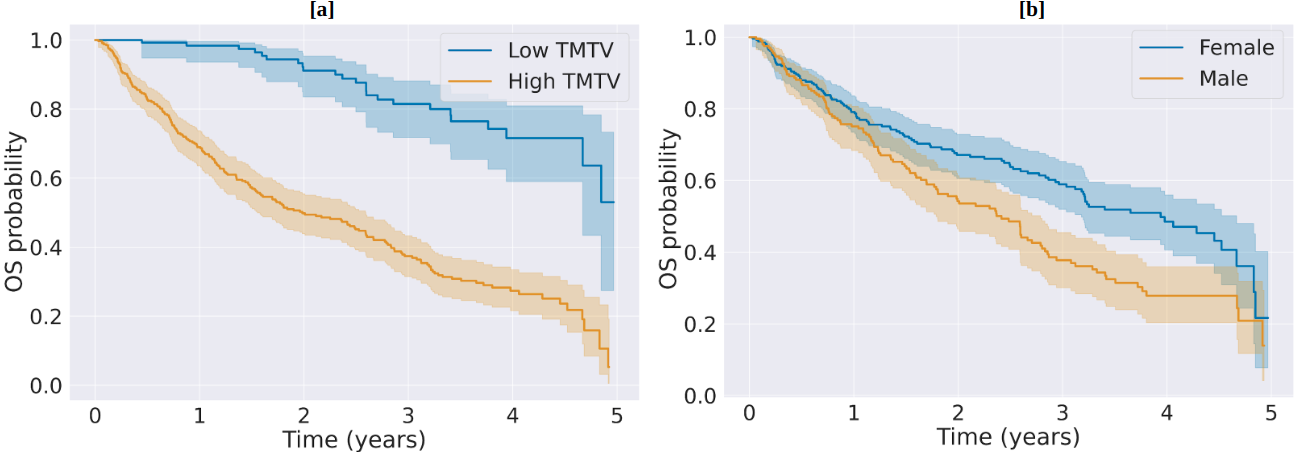}
  \caption{\small Kaplan-Meier curves illustrating overall survival (OS) probabilities as a function of time on the test set for the following comparisons: [a] low vs high TMTV based on median cut-off value, [b] female vs male.}
\end{figure}

\newpage
\section{Pycox Deep Learning Models: Training and Feature Details} \label{appendix:clin_IDP}

We evaluated two deep learning-based survival models DeepHit and DeepSurv, both implemented using the \texttt{pycox} package. More details can be found in the official GitHub repository: \url{https://github.com/havakv/pycox}. Each model was trained for 100 epochs using 5-fold cross-validation with clinical data as input. Predictions on the test set were obtained by aggregating the results from each fold at every time point.

\subsection*{Model descriptions}
\begin{itemize}
    \item \textbf{DeepHit:} A deep neural network that models the full discrete-time survival distribution. It can accommodate non-proportional hazards and competing risks, offering flexibility for complex datasets. In this study, we used the single-event implementation provided by the \texttt{pycox} package.
    \item \textbf{DeepSurv:} A deep learning extension of the Cox proportional hazards model that captures nonlinear covariate effects while maintaining the proportional hazards assumption.
\end{itemize}

\subsection*{Features used}
\begin{itemize}
    \item \textbf{Age}
    \item \textbf{Sex}
    \item \textbf{TMTV}
    \item \textbf{Dmax $>$ Q3}: Binary indicator of whether Dmax (maximum distance between lesions on PET) exceeds Q3.
    \item \textbf{Lesion Count $>$ Median}: Binary indicator of whether the lesion count exceeds the median.
    \item \textbf{T-Staging}: The T-stage describes the size and extent of the primary tumor. To incorporate T-stage data into the model, we used one-hot encoding for categorical T-stage values. For cases with missing T-stage data, we introduced an additional binary column to explicitly indicate the absence of the value (NaN), ensuring that missingness was preserved.
    
    \item \textbf{N-Staging}: N-stage describes the spread of the cancer to the regional lymph nodes. To incorporate N-stage data into the model, we used one-hot encoding for categorical N-stage values. For cases with missing N-stage data, we introduced an additional binary column to explicitly indicate the absence of the value (NaN), ensuring that missingness was preserved.
\end{itemize}

\end{document}